\RequirePackage[l2tabu,orthodox]{nag}
\documentclass[letterpaper, 10 pt, conference]{ieeeconf} 	% Openright aabner kapitler paa hoejresider (openany begge)

\usepackage[margin=0.75in]{geometry}
\usepackage[utf8]{inputenc}					% Input-indkodning af tegnsaet (UTF8)
\usepackage[british]{babel}					% Dokumentets sprog
\usepackage[T1]{fontenc}					% Output-indkodning af tegnsaet (T1)
\usepackage{hyperref}

\usepackage{amsmath,amssymb} 
\usepackage[noend]{algpseudocode}		% Avancerede matematik-udvidelser
\usepackage{mathtools}						% Andre matematik- og tegnudvidelser
\usepackage{setspace}

\usepackage[backend=bibtex,maxbibnames=99]{biblatex}
\usepackage{tikz}
\usetikzlibrary{patterns}

\usepackage[amsmath,thmmarks,hyperref]{ntheorem}
\theoremstyle{marginbreak}
\theoremseparator{:}
\global
\global

\newtheorem{thm}{Theorem}
\newtheorem{defi}[thm]{Definition}

\newtheorem{prop}[thm]{Proposition}
\theorembodyfont{}

\theoremsymbol{\ensuremath{\blacktriangleleft}}

\theoremstyle{nonumberbreak}
\theoremsymbol{}

\theoremsymbol{\tikz \fill[black] (0,0) rectangle (1.2ex,1.2ex);}

\usepackage{empheq}
\usepackage{pdflscape}

\newcommand{\vek}[1]{\boldsymbol{\mathbf{#1}}}

\usepackage{newfloat}
\DeclareFloatingEnvironment[
    fileext=los,
    listname={List of Schemes},
    name=Protocol,
    placement=tbhp,
]
{protocol}

\newcommand{\sbline}{\\[.5\normalbaselineskip]}% small 

\title{Secure PAC Bayesian Regression via \\Real Shamir Secret Sharing}
\author{
     Jaron Skovsted Gundersen\thanks{
     jaron@es.aau.dk
 },\;
     Bulut Kuskonmaz\thanks{bku@es.aau.dk},\;
     Rafael Wisniewski\thanks{raf@es.aau.dk}
 }

\date{  Aalborg University\\ 
        Department of Electronic System\\\vspace{10pt}
        \today}

\begin{document}
  \maketitle
\begin{abstract}
A common approach of system identification and machine learning is to generate a model by using training data to predict the test data instances as accurately as possible. Nonetheless, concerns about data privacy are increasingly raised, but not always addressed. We present a secure protocol for learning a linear model relying on a recently described technique called real number secret sharing. We take as our starting point the PAC Bayesian bounds and deduce a closed form for the model parameters which depends on the data and the prior from the PAC Bayesian bounds. To obtain the model parameters one needs to solve a linear system. However, we consider the situation where several parties hold different data instances and they are not willing to give up the privacy of the data. Hence, we suggest to use real number secret sharing and multiparty computation to share the data and solve the linear regression with a secure distributed Gaussian elimination protocol such that privacy of the data is preserved. The benefit of using secret sharing directly on real numbers is reflected in the simplicity of the protocols and the number of rounds needed. However, this comes with the drawback that a share might leak a small amount of information, but in our analysis we argue that the leakage is small. 
\end{abstract}

\section{Introduction}
\label{sec:intro}
The main purpose of system identification and machine learning is to find a model based on data such that the model generalizes to instances outside the data set. The more data we possess the better is the model. However, the data may be sensitive and hence we are not allowed to share the data with others. %This could be the case if the data regards the price of houses, health, or bank account information. 
In this paper, we consider the case where several parties hold a data set and each party wants to learn a model trained on the data in all the data sets. We strive to learn the model without giving up the privacy of the data. One way to determine a model is to rely on the PAC Bayesian framework, which can be used if some prior knowledge of the model is available. There are a lot of PAC Bayesian bounds in the literature which assess the generalization error with a high probability, see for instance \cite{AlquierBound, CatoniBound, LangfordBound, McAllesterBound}. We use the PAC Bayesian framework with a normally distributed prior to obtain a linear model estimation. In this setup, we get a closed form solution for the estimation. 

For the purpose of privacy, techniques from secure multiparty computation (MPC) are useful. In \cite{Ridge_reg}, they apply homomorphic encryption and Yao garbled circuits to achieve the secure ridge regression as a secure machine learning algorithm. In \cite{Lin_Reg_dist} MPC is used to securely compute linear regression without prior and where the data is split between two parties. Another study uses encrypted data for their classification phases as they call it privacy-preserving classification \cite{bost2014machine}. Originally, and in these papers, MPC is carried out over a finite field, but recent research considers MPC over real numbers \cite{Katrine_Shamir} and we take this approach. However, this is a fairly new approach and in \cite{Katrine_Shamir} the privacy is only shown for a single operation. In this paper we want to argue that the privacy is retained even when several secure computations are carried out sequentially. 

With the assumption that data is split between different parties, we show how one can use real number secret sharing and MPC techniques to compute the model without violating privacy. To argue privacy, we consider the notion of conditional mutual information $I(X;V|Y)$. In our setup $Y$ is everything that an adversary will learn in an ideal world (the data of the corrupted parties and the model), $X$ is what we want to hide (the data of the honest parties), and $V$ is the additional information an adversary gets by using the protocol. Thus, $I(X;V|Y)$ describes how much more information the adversary learns about the honest parties data by using the protocol compared to an ideal version where only the model is given to them. Ideally this quantity should be zero which is equivalent to that $X$ and $V$ are conditionally independent given $Y$. Since $I(X;V|Y)$ is hard to determine in our case, due to the high dimensions of $X,V,Y$, we will use a conditionally independent test to argue privacy. Conditionally independent tests are a line of research of its own. More recently, kernel based and neural network based tests have been developed which is argued to be applicable in the high dimensional cases \cite{Kernel_test_independence,GAN_independence}. 

\section{Preliminaries}
\subsection{Notation}\label{Sec_Notation}
Let $\mathcal{D}=\mathcal{D}_1\cup\mathcal{D}_2\cup\cdots\cup \mathcal{D}_n$ be a split data set and assume that party $i$ holds $$\mathcal{D}_i=\{(\vek{x}_{i1},y_{i1}),(\vek{x}_{i2},y_{i2}),\ldots,(\vek{x}_{in_i},y_{in_i})\}.$$
Furthermore, we denote by $N=\sum_{i=1}^n n_i=|\mathcal{D}|$ and assume that $(\vek{x}_{ij},y_{ij})\in \mathbb{R}^d\times \mathbb{R}$ where $d$ is the dimension of data. 

Assume that a model $f$ is given. For a data point $(\vek{x},y)\in\mathcal{D}$, the prediction of $\vek{x}$ is $f(\vek{x})$ and the squared loss is denoted by $\ell(f(\vek{x}),y)=(f(\vek{x})-y)^2$. The empirical loss is
\begin{align}\label{eq:Loss_func}
    \mathcal{L}_N(f)=\frac{1}{N}\sum_{i=1}^n\sum_{j=1}^{n_i}\ell(f(\vek{x}_{ij}),y_{ij}),   
\end{align}
and the generalization error is the expected value of the loss function, i.e., $\mathcal{L}(f)=E_{(\vek{x},y)} [\ell(f(\vek{x}),y))]$, where the expectation is taken over the joint distribution of $\vek{x}$ and $y$. 

\subsection{Multiparty computation}
MPC deals with the situation where there are $n$ parties, and each has some inputs to a function. The parties would like to learn the output of the function without revealing their inputs to the other parties. An MPC protocol should make it possible for the parties to obtain this even in the presence of an adversary corrupting some of the parties. In this paper, we consider a passive corrupt adversary, meaning that the adversary can see everything the corrupted parties are able to see but they cannot deviate from the protocol description. To prove that a protocol is secure we usually compare what harm an adversary can do in an execution of the protocol to what harm an adversary can do in an idealized version. Since, we are considering a passive adversary the only ``harm'' is that the adversary learns the information it was not supposed to learn. However, this can also be formulated through mutual information\footnote{For more information about mutual information we refer to \cite{CoverBook}.}. Let $X$ and $Y$ be random variables then the mutual information can be defined 
\begin{align*}
    I(X;Y)   &= h(X)-h(X|Y)
\end{align*}
where $h(X)$ is the entropy of $X$ and $h(X|Y)$ is the conditional entropy.\footnote{Mutual information can also equivalently be defined through the Kullback-Leibler divergence where the joint distribution of $X$ and $Y$ is compared to the product of the marginals.} With this definition we use the same definition as in \cite{LGHC21} to define privacy of a protocol.
\begin{defi}[Perfect Privacy]
Let $\mathrm{View}$ be anything the adversary sees through a protocol computing $f(x_1,\ldots x_n)=y$ and let $\mathcal{A}$ and $\mathcal{H}$ be the indices of the inputs held by corrupted and honest nodes. Then we say that the protocol obtains perfect privacy if
\begin{equation*}
    I(\{x_i\}_{i\in \mathcal{H}};\mathrm{View})=I(\{x_i\}_{i\in \mathcal{H}};\{y\}\cup\{x_j\}_{j\in \mathcal{A}}).
\end{equation*}
\end{defi}
When considering continuous random variables, $ I(\{x_i\}_{i\in \mathcal{H}};\mathrm{View})$ might be very difficult to compute since we are rarely able to determine the distribution on $\mathrm{View}$ and it will often have a very high dimension.\footnote{This is for instance mentioned in \cite{KGW22} where alternatives to the mutual information is suggested by using another measure than the KL-divergence. However, these suggestions have also computational problems when the dimension increases.} 

Since the mutual information is difficult to compute in our setup, for instance because of the random variables have high dimensions, we take another approach. Note that $\mathrm{View}= V\cup \{y\}\cup\{x_j\}_{j\in \mathcal{A}}$ where $V$ includes all the randomness chosen by the adversary and all the messages the adversary receives. Since we do not obtain perfect privacy we want to show that $V$ does not reveal much information about $\{x_i\}_{i\in \mathcal{H}}$. That is we want 
\begin{align}
    &I(\{x_i\}_{i\in \mathcal{H}};\mathrm{View})-I(\{x_i\}_{i\in \mathcal{H}};\{y\}\cup\{x_j\}_{j\in \mathcal{A}})\nonumber \\&= I(\{x_i\}_{i\in \mathcal{H}};V|\{y\}\cup\{x_j\}_{j\in \mathcal{A}}).\label{eq:Cond_MI}
\end{align}
to be small. If $\{x_i\}_{i\in \mathcal{H}}$ and $V$ are conditionally independent conditioned on $\{y\}\cup\{x_j\}_{j\in \mathcal{A}}$ this conditional mutual information is zero. Thus in our privacy analysis we apply a conditional independence test to show that if we apply our distributed algorithm on a given data set with large enough security parameters we cannot reject the null hypothesis that $\{x_i\}_{i\in \mathcal{H}}$ and $V$ are conditionally independent given $\{y\}\cup\{x_j\}_{j\in \mathcal{A}}$. And hence $I(\{x_i\}_{i\in \mathcal{H}};V|\{y\}\cup\{x_j\}_{j\in \mathcal{A}})$ must be close to zero. We apply the test from \cite{GAN_independence}. 

Many MPC protocols are secret sharing-based meaning that each party starts by sending a share of its input to all the other parties. The share itself does not reveal information and in fact we say that the secret sharing scheme has privacy threshold $t$, if $t$ shares do not reveal anything about the secret. However, if more than $t$ shares are collected we should be able to reconstruct the secret. MPC can use properties for certain secret sharing schemes allowing us to do computations on the shares. From these computations, we end up with the parties having a share for the output. By broadcasting these shares, they can by the reconstruction property in the secret sharing scheme compute the output. For a comprehensive introduction to secret sharing and MPC, we refer the reader to \cite{cramer_damgard_nielsen_2015}. 

Typically MPC is carried out over a finite field and therefore it is only shown how to securely carry out summation and multiplication. We would like to compute over the real numbers and therefore we need to be able to carry out other computations as well. Some works embed secure computations on real numbers into finite field operations by representing the real numbers as fixed or floating point numbers. This is for instance the case in the papers \cite{aliasgari2013secure, Fixed-point2, MPC_Fixed, Dimitrov_etal}. One of the main challenges this approach faces is to carry out real number divisions in a secure way as division in a finite field and integer division are completely different. As an example the division $5/2$ is $8$ in the field $\mathbb{F}_{11}$ but it is $2$ as integer division. Hence one needs to introduce other tools to carry out integer division in the finite field. For instance, \cite{aliasgari2013secure, MPC_Fixed} relies on a iterative algorithm for computing the division. But this comes with the drawback that they need $17$ communication rounds before they obtain the result if they use bitlength $32$ for their floating points or fixed points. Instead of transforming the real number computations into finite field operations we take another approach and consider the computations directly in the real numbers. 

To obtain simpler protocols for the secure arithmetic over the real numbers, we consider real number secret sharing introduced in \cite{Katrine_Shamir}. This concept is inspired by Shamir's secret sharing over finite fields \cite{Shamir}, and considering this in the real numbers comes with the drawback that a share might reveal a small amount of information. To secretly share a value $s$ in a finite field $\in \mathbb{F}$ using Shamir's scheme, we choose $c_i\in \mathbb{F}$, uniformly at random for $i=1,2,\ldots,t$ and compute the polynomial 
\begin{equation}\label{eq:Shamir_pol_field}
    f_s(x)=s+c_1x+c_2x^2+\cdots c_tx^t.
\end{equation} 
The shares are evaluation points on this polynomial, meaning that for an $\alpha\in \mathbb{F}$ a share is $f_s(\alpha)$. The privacy relies on the uniform distribution in $\mathbb{F}$ combined with Lagrange interpolation arguments stating that $t+1$ points uniquely determines a degree-$t$ polynomial. Hence, we also have the reconstruction property.

\subsection{MPC based on real number secret sharing}\label{sec:MPC_real}
We give a brief introduction to the ideas from \cite{Katrine_Shamir}. One of the obstacles they had was that if the shares were constructed from the polynomial in \eqref{eq:Shamir_pol_field} the information leakage from the shares depends on how close the evaluation point is to $0$. Furthermore, there is no uniform distribution on the real axis. %meaning that we do not need to encode the numbers into some finite field for secure computations.

To circumvent the first obstacle the polynomial $f_s$ is constructed in another way by using the Lagrange polynomial as a basis for all polynomial of degree at most $t$ instead of the basis $\{1, x, \ldots x^t\}$. To be a bit more formal, let $\mathcal{P}=\{\alpha_i\}_{i=1,\ldots,n}$. Each time a secret $s$ needs to be shared a polynomial is constructed in the following way. Choose randomly a subset $\mathcal{P}_s'\subset \mathcal{P}$ of size $t$ and let $\mathcal{P}_s=\mathcal{P}_s'\cup\{0\}$. Consider the $t+1$ points $\mathcal{T}=\{(\alpha_i,\beta_i)\}_{i \in \mathcal{P}_s}$, where we use $(\alpha_0,\beta_0)=(0,s)$. The $\beta_i$'s for $i>0$ are chosen randomly according to a normal distribution with variance $\sigma_\beta^2$. Define the Lagrange basis polynomials as $L_i(x)=\prod_{j\in \mathcal{P}_s\setminus\{i\}}\frac{x-\alpha_j}{\alpha_i-\alpha_j}$. Then we define 
\begin{align}
    f_s(x)=\sum_{i\in \mathcal{P}_s}\beta_i L_i(x).
\end{align}
Since $L_i(\alpha_j)=0$ when $i\in\mathcal{T}\setminus \{i\}$ and $L_i(\alpha_i)=1$, we obtain that $f_s(x)$ is a degree-$t$ polynomial passing through the points in $\mathcal{T}$. The shares are evaluation points for this polynomial in the elements in $\mathcal{P}$ meaning that a share of the secret $s$ to the $i$'th party is $f_s(\alpha_i)$. To denote that $s$ is secretly shared we write $[s]$, which represents the vector $(f_s(\alpha_1),f_s(\alpha_2),\ldots,f_s(\alpha_n))$ where the $i$'th party knows the $i$'th entry.

We end this section with a description of how to compute on the shares. In \cite{Katrine_Shamir}, protocols for securely computing sums, multiplications, and divisions over the real numbers are presented which are generalizations of protocols for finite fields described in \cite{cramer_damgard_nielsen_2015}. However, we take fairly different approaches than in \cite{Katrine_Shamir} for multiplications and inversions. The reason for changing the secure multiplication is due to the fact that we can lower the amount of information sent if we assume that $t<\frac{n}{2}$ and take advantage of some of the properties the Shamir secret sharing possess. The reason we change the secure inversion is to correct a small flaw in the privacy analysis in \cite{Katrine_Shamir}. Furthermore, we mention that privacy analysis for a single operation (one multiplication, one inversion, etc.) is provided in \cite{Katrine_Shamir}, and we will do the same for our secure operations, but how to analyse the privacy of an algorithm as the one we will present in Section \ref{sec:prot} is not described. In this paper, we suggest to use a conditional independence test for that. 

To carry out a secure sum, the parties locally add up the shares for the two secrets. Specifically, assume that 
\begin{align}
\begin{aligned}
    &[x]=(f_x(\alpha_1),f_x(\alpha_2),\ldots,f_x(\alpha_n))\\ &[y]=(f_y(\alpha_1),f_y(\alpha_2),\ldots,f_y(\alpha_n))
\end{aligned}
\end{align}
then by $[x]+[y]$ we mean that the parties locally sum up their shares. We remark that $f_x(0)=x$ and $f_y(0)=y$ and the sum of the shares represents a degree-$t$ polynomial $f_{x+y}$ which evaluates to $x+y$ at zero. Hence, we can write $[x]+[y]=[x+y]$. Similarly, we can add a known value to a secret shared value by letting the parties add the value to their share and we can multiply a known value to a secret shared value by letting each party multiply its share by that given value. Thus, we also use the notation $a+[x]=[a+x]$ and $a\cdot [x]=[a\cdot x]$ to represent these computations. 

We remark that all operations so far have not required any communication between the parties if $[x]$ and $[y]$ are obtained beforehand. However, for the multiplication and division protocols we require some communication. 

For the secure multiplication we assume that $t<\frac{n}{2}$ which allows us to carry out a single multiplication (and linear combinations of such values) locally. Notice that $(f_x(\alpha_1)f_y(\alpha_1),f_x(\alpha_2)f_y(\alpha_2),\ldots,f_x(\alpha_n)f_y(\alpha_n))$ corresponds to evaluations of a degree-$2t$ polynomial with constant term $xy$. Thus a multiplication can also be carried out locally and we write $[x]\cdot[y]=[xy]_{2t}$. Linear combination of $[\cdot]$ and $[\cdot]_{2t}$ shares can be carried out locally as well but if $[\cdot]_{2t}$ shares needs to be multiplied by another secret value we need to ``refresh'' the shares. To refresh $[x]_{2t}$ we need some preprocessed shares $([r]_{2t},[r])$, where $r$ is normal distributed with variance $\sigma_r^2$. Then, $[x]_{2t}-[r]_{2t}$ can be computed and opened, and afterwards $x-r+[r]=[x]$. When we say a value is opened it means that all parties reveal their share of the value and the value can be reconstructed. In practice, and in our analysis later on, we implement this as all parties are sending their share to the first party who reconstruct the value and send it back. 

For the multiplication, we converted $[x]_{2t}$ to a degree-$t$ sharing. In the remaining we skip the subscript $2t$ for simplicity but we note that one has to take care of the conversions in the algorithm. We show that revealing $x-r$ do not give much information about $x$ if $\sigma_r^2$ is high. 
\begin{prop}\label{prop:mult_leak}
    Let $x$ and $r$ be independent and let $r\sim \mathcal{N}(\mu_r,\sigma_r^2)$ and $\sigma_x^2=\mathrm{Var}(x)$. Then
    \begin{align*}
        I(x;x-r)\leq \frac {1}{2}\log\left( 1+\frac{\sigma_x^2}{\sigma_r^2}\right).
    \end{align*}
\end{prop}
\begin{proof}
We use the definition of mutual information 
    \begin{align*}
        I(x;x-r)=h(x-r)-h(x-r|x).
    \end{align*}
    The last term is the entropy of a normal distribution and the first term can be upper bounded by the entropy of such a distribution since for a giving variance the entropy of a normal distribution is maximal. Hence, we have
    \begin{align*}
        I(x;x-r) &\leq\frac {1}{2}\log \left(2\pi e(\sigma_x^2+\sigma_r^2 )\right)-\frac {1}{2}\log \left(2\pi e \sigma_r^2 \right)\\
        &=\frac {1}{2}\log\left( 1+\frac{\sigma_x^2}{\sigma_r^2}\right).
    \end{align*}
\end{proof}

At last, we describe how to invert real numbers securely. Here, we need some preprocessed $[be^{r}]$, where $r$ is normal distributed with variance $\sigma_{e^r}^2$ and $b$ is chosen uniform randomly from $\{-1,1\}$. First $[x]\cdot [be^r]=[xbe^r]_{2t}$ can be computed locally as described above and opened towards the parties. After $xbe^r$ is opened we can compute $(xbe^r)^{-1}[be^r]=[x^{-1}]$. We remark that this secure inversion is a bit different than in \cite{Katrine_Shamir} since we assume another distribution on the obfuscation. Therefore, we prove that a single inversion is not leaking much information.

I HAVE CHANGED THIS PART A BIT. WE MIGHT NEED TO DO SOMETHING IN EXPERIMENTS.

%At last, we describe how to invert real numbers securely. Here we need some preprocessed $([r],[e^{r}],[r'],[e^{r'}])$, where $r$ and $r'$ are normal distributed with variance $\sigma_r^2$. First $[x]\cdot [e^r]=[xe^r]_{2t}$ can be computed locally as described above and opened towards the parties. We remark that this is actually revealing the sign of $x$ since $\mathrm{sgn}(x)=\mathrm{sgn}(xe^r)$. However, with respect to mutual information the sign is not giving much information as we show below. After $xe^r$ is opened we can compute $\log(\mathrm{sgn}(x)xe^r)-[r]=[\log(\mathrm{sgn}(x)x)]$. Now compute $-[\log(\mathrm{sgn}(x)x)]-[r']$ and open this. At last $\mathrm{sgn}(x)e^{-\log(\mathrm{sgn}(x)x)-r'}\cdot [e^{r'}]=[\frac{1}{x}]$. Since secure inversion method is not presented in \cite{Katrine_Shamir} we prove that a single inversion is not leaking much information
\begin{prop}\label{prop:inversion_leak}
    Let $x$ and $r$ be independent and let $r\sim \mathcal{N}(\mu_{e^r},\sigma_{e^r}^2)$ and $\sigma_z^2=\mathrm{Var}(\log|x|)$. Then
    \begin{align*}
        I(x;xbe^r)\leq \frac {1}{2}\log\left( 1+\frac{\sigma_z^2}{\sigma_{e^r}^2}\right).
    \end{align*}
\end{prop}
\begin{proof}
We use that $I(x;y,z)= I(x;g(y,z))$ when $g$ bijective \cite{EstMI}, meaning that
    \begin{align*}
        I(x;xbe^r)=I(x;\mathrm{sgn}(xb),\log|x|+r).
    \end{align*}
    Note that $\mathrm{sgn}(xb)$ is $-1$ with probability $\frac{1}{2}$ and $1$ with probability $\frac{1}{2}$ no matter the distribution on $x$. Hence, $\mathrm{sgn}(xb)$ is independent on $x$ and $\log|x|+r$ and we can remove this term. Using the definition of mutual information we obtain
    \begin{align*}
        I(x;\log|x|+r)=h(\log|x|+r)-h(\log|x|+r|x).
    \end{align*}
    The last term is the entropy of a normal distribution and the first can be upper bounded as in the proof of Proposition \ref{prop:mult_leak}.
    \begin{align*}
        I(x;xbe^r)&\leq \frac {1}{2}\log \left(2\pi e(\sigma_z^2+\sigma_{e^r}^2 )\right)-\frac {1}{2}\log \left(2\pi e \sigma_{e^r}^2 \right)\\
        &=\frac {1}{2}\log\left( 1+\frac{\sigma_z^2}{\sigma_{e^r}^2}\right).
    \end{align*}
\end{proof}
Remark that in both Propostion \ref{prop:mult_leak} and \ref{prop:inversion_leak} we can control the leakage of $x$ by varying the variance of $r$.

%We refer to \cite{Katrine_Shamir}, for details about the subprotocols for computing $[x]\cdot [y] = [x\cdot y]$ and $1/[x]=[1/x]$, but mention here that the former make use of the already described protocols above and a preprocessed so-called Beaver triple $([a],[b],[c])$, where $a$ and $b$ are random and $c=ab$. The latter subprotocol make also use of the above described protocols and a preprocessed random sharing $[r]$. Both of these subprotocols require that some values are opened. This means that for a sharing $[x]$, we reveal the value $x$ to the parties. This can be done by broadcasting the shares to the other parties. Hence, both of the subprotocols require some communication between the parties.

We will slightly abuse the notation and write $[X]$ and $[\vek{x}]$ for a matrix $X$ and vector $\vek{x}$. This means that all entries should be secret shared and the computations are entry-wise.

\subsection{PAC Bayesian and linear regression}
Section \ref{Sec_Notation} considered the loss of a given model. In this section we want to determine the model from a Bayesian point of view, and hence we assume that $f$ is random. 

We consider the generalization error from \cite{AlquierBound} and \cite{germain2016} which states that with probability more than $1-\delta$ the following inequality is satisfied
\begin{align}\label{eq:PACBayes}
    \begin{aligned}
	E_{f\sim \hat{\rho}}&\mathcal{L}(f)\leq E_{f\sim \hat{\rho}}\mathcal{{L}}_{N}(f)\\& +\frac{1}{\lambda}\left(KL(\hat{\rho},\pi)+\ln\frac{1}{\delta}+\Psi_{\ell,\pi}(\lambda,N)\right),
	\end{aligned}
\end{align}
where $\pi$ is the prior distribution of $f$, $\hat{\rho}$ the posterior distribution, $\lambda$ is a tuning parameter which we return to, and $\Psi_{\ell,\pi}(\lambda,N)=\ln E_{f\sim\pi}E_{(\vek{x},y)}e^{\lambda(\mathcal{L}(f)-\mathcal{L}_{N}(f))}$, where $\mathcal{L}(f)$ and $\mathcal{L}_{N}(f))$ are as in \eqref{eq:Loss_func} and below this equation.

The left-hand side in \eqref{eq:PACBayes} is the expected generalization error, so we upper bound this with a high probability. For a given data set, tuning parameters and prior, in order to minimize the right-hand side we need a posterior $\hat{\rho}$ minimizing $E_{f\sim \hat{\rho}}\mathcal{{L}}_{{N}}(f)+\frac{1}{\lambda}KL(\hat{\rho},\pi)$ since these are the only terms depending on $\hat{\rho}$. The Gibbs posterior
\begin{align}\label{eq:Gibbs}
	\hat{\rho}^*(f)=\frac{1}{E_{f\sim \pi}e^{-\lambda {\mathcal{L}}_{N}(f)}}\pi(f)e^{-\lambda {\mathcal{L}}_{N}(f)}
\end{align}
minimizes this \cite{AlquierBound}. To choose $f$ from $\hat{\rho}^*(f)$, we set $f_*=\mathrm{argmax}_f ~\hat{\rho}^*(f)$. Hence, w.r.t. $f$, we need to minimize
\begin{align}\label{eq:WhatToMin}
	\lambda {\mathcal{L}}_{N}(f)-\ln \pi(f).
\end{align}
We define our model space $\mathcal{F}$ to consists of linear models $f(\vek{x})=\vek{w}\cdot\vek{x}$ for some $\vek{w}\in \mathbb{R}^d$. We put some restrictions on our prior. Since $f$ is uniquely determined by $\vek{w}$ we define the prior on $\vek{w}$. We assume a normal distributed prior $\pi(\vek{w})=\mathcal{N}(\vek{\mu}_w;\Sigma_w)$, where $\Sigma_w$ is positive definite. With these assumptions \eqref{eq:WhatToMin} reduces to that we have to minimize
\begin{align*}
	\frac{\lambda}{N} \sum_{i=1}^n\sum_{j=1}^{n_i}(\vek{w}\cdot\vek{x}_{ij}-y_{ij})^2-\frac{1}{2}(\vek{w}-\vek{\mu}_w)^T\Sigma_w^{-1}(\vek{w}-\vek{\mu}_w)
\end{align*}    
with respect to $\vek{w}$. This is a convex function, so differentiating and setting equal to $\vek{0}$ yields the minimum. Hence, to minimize it, we solve the following equation.
\begin{align}\label{eq:Optimal}
\scriptsize{
\begin{aligned}
	  \left(\frac{2\lambda}{N}\sum_{i=1}^n\sum_{j=1}^{n_i}\vek{x}_{ij}\vek{x}_{ij}^T+\Sigma_w^{-1}\right)\vek{w}=\frac{2\lambda}{N}\sum_{i=1}^n\sum_{j=1}^{n_i} \vek{x}_{ij}y_{ij}+\Sigma_w^{-1}\vek{\mu}_w.
\end{aligned}	 }
\end{align}
Setting $\lambda=0$ we obtain $\vek{w}=\vek{\mu}_w$ meaning that the posterior relies fully on the prior. Opposite, letting $\lambda$ tend to infinity the posterior relies fully on the data.
%$$\vek{w}=\left(\sum_{i=1}^n\sum_{j=1}^{n_i}\vek{x}_{ij}\vek{x}_{ij}^T\right)^{-1}\sum_{i=1}^n\sum_{j=1}^{n_i} \vek{x}_{ij}y_{ij}$$ 

\hspace{1pt}

\section{Secure distributed regression using PAC Bayes approach}\label{sec:prot}
Consider the distributed part, where the parties want to find the solution to \eqref{eq:Optimal} without revealing the datapoints in $\mathcal{D}_i$. Note that the parties can compute the inner sums locally,
\begin{align*}
	{X}_i=\sum_{j=1}^{n_i} \vek{x}_{ij}\vek{x}_{ij}^T, \quad {\vek{z}}_i=\sum_{j=1}^{n_i} \vek{x}_{ij}y_{ij},
\end{align*}
implying that \eqref{eq:Optimal} can be rewritten as
\begin{align*}
	\left(\frac{2\lambda}{N}\sum_{i=1}^n X_i+\Sigma_w^{-1}\right)\vek{w}=\frac{2\lambda}{N}\sum_{i=1}^n\vek{z}_i+\Sigma_w^{-1}\vek{\mu}_w.
\end{align*}
This is similar to the local part in \cite{Ridge_reg}, but we remark that the setup in their paper is different from ours since the computation takes place at some servers.

We need now a secure way to compute the matrix $A=\frac{2\lambda}{N}\sum_{i=1}^n X_i+\Sigma_w^{-1}$ and the vector $\vek{b}=\frac{2\lambda}{N}\sum_{i=1}^n\vek{z}_i+\Sigma_w^{-1}\vek{\mu}_w$. Furthermore, we need a way to solve $A\vek{w}=\vek{b}$ in a secure way without revealing $A$ and $\vek{b}$. We start with the $i$'th party locally compute $X_i$ and $\vek{z}_i$ and secretly share the entries between all parties. Then by using the computations on shares described in Section \ref{sec:MPC_real} the parties can compute $[\frac{2\lambda}{N}\sum_{i=1}^n X_i+\Sigma_w^{-1}]$ and $[\frac{2\lambda}{N}\sum_{i=1}^n\vek{z}_i+\Sigma_w^{-1}\vek{\mu}_w]$. Now we have $A$ and $\vek{b}$ secretly shared and we need a way to secretly solve the system $A\vek{w}=\vek{b}$.

Notice that $X_i$ is a positive semidefinite matrix and hence $A$ is positive definite due to the assumption on $\Sigma_w$. Thus, the matrix is invertible and the system in \eqref{eq:Optimal} has a unique solution. Furthermore, since $A$ is symmetric and positive definite we can do Gaussian elimination without pivoting. This means that we never have to interchange rows when row-reducing since the diagonal entries will always be nonzero. After we obtain an upper-triangular matrix, we solve the system with backward substitution. The protocol can be found in Protocol \ref{prot:regres}.

\begin{protocol}[t]
\noindent\rule[-10pt]{\linewidth}{.8pt}\vspace{1pt}
    \textit{Input:} $P_i$ holds $\mathcal{D}_i=\{(\vek{x}_{ij}, y_{ij})\mid j=1,2,\ldots, n_i\}$, where $N=\sum_{i=1}^n n_i$ is known. The tuning parameter $\lambda$, $\vek{\mu}_w$, and a positive definite $\Sigma_w$ determining the prior on $\vek{w}$ is fixed.
\sbline
\textit{Output:} The parties should learn $\vek{w}$ satisfying \eqref{eq:Optimal}.
\sbline
\textit{The protocol:}
\begin{enumerate}
    	\item $P_i$ sets ${X}_i=\sum_{j=1}^{n_i} \vek{x}_{ij}\vek{x}_{ij}^T$ and ${\vek{z}}_i=\sum_{j=1}^{n_i} \vek{x}_{ij}y_{ij}$
		\item $P_i$ send shares of $X_i$ and $\vek{z}_i$ to the other parties.
		\item Using the secure protocols for summation, multiplication by scalar, and adding scalar, the parties compute 
		\begin{equation*}
		\begin{aligned}
		    [A]&=\frac{2\lambda}{N}\sum_{i=1}^n [X_i]+\Sigma_w^{-1}\\
                [\vek{b}]&=\frac{2\lambda}{N}\sum_{i=1}^n[\vek{z}_i]+\Sigma_w^{-1}\vek{\mu}_w
		\end{aligned}
		\end{equation*}
		%\item Solve $A\vek{w}=\vek{b}$ securely by giving $[A]$ and $[\vek{b}]$ as input to Protocol \ref{prot:Gauss}.
		\item Denote by $[C]=\left[[A]~|~[\vek{b}]\right]$ the $d\times (d+1)$ total matrix with shared entries $[c_{ij}]$
    \item
        \begin{algorithmic}
        \For{$k=1\ldots d-1$}
            \State $[\mathrm{inv}_k]=\frac{1}{[c_{kk}]}$
			\For{$i=k+1\ldots d$}
			    \State $[\mathrm{frac}_{ik}] = [c_{ik}]\cdot [\mathrm{inv}_k]$
                \State $[c_{ik}]=[0]$
				\For{$j=k+1\ldots d+1$}
				    \State  $[c_{ij}]=[c_{ij}]-[\mathrm{frac}_{ik}]\cdot[c_{kj}]$
				\EndFor
			\EndFor
		\EndFor
        \end{algorithmic}
    $C$ is now an upper triangular matrix with nonzero elements on the diagonal
    
    \item 
    \begin{algorithmic}
        \For{$i=0\ldots d-1$}
			\State $[w_{d-i}]=\frac{[c_{d-i,d+1}]-\sum_{j=1}^i [c_{d-i,d-i+j}]w_{d-i+j}}{[c_{d-i,d-i}]}$
			\State Open $w_{d-i}$ to all parties
		\EndFor
    \end{algorithmic}
\end{enumerate}
\vspace{-5pt}
\caption{Secure Regression with prior from Gaussian elimination}\label{prot:regres}
\noindent\rule{\linewidth}{.8pt}\vspace{-\baselineskip}%
\end{protocol}

%In the following, we describe these two methods for solving the linear system securely and compare the two methods afterwards. One way to solve the system is to compute the inverse of $A$ securely and multiply it with $\vek{b}$ in a secure way as described in the previous section. We remark, that $[e^R]\cdot[A]$ is matrix multiplication but since we have subprotocols for addition and multiplication we can also carry out matrix multiplication using these subprotocols. Furthermore, we remark that with the assumption that $t<\frac{n}{2}$, we can compute $[e^RA]_{2t}$ without interaction. The complete protocol for this Secure Inverse Method (SIM) can be found in Protocol \ref{prot:Inverse}.

%\begin{protocol}[t]
%\noindent\rule[-10pt]{\linewidth}{.8pt}\vspace{1pt}
%\textit{Input:} $[A]$ and $[\vek{b}]$
%\sbline
%\textit{Output:} $\vek{w}$ satisfying $A\vek{w}=\vek{b}$.
%\begin{enumerate}
%    \item Compute $[e^R]\cdot [A]=[e^RA]_{2t}$ locally by making multiplactions of the shares and linear combinations of those locally. Afterwards, open the matrix such that each party learns $e^RA$
%    \item Compute $(e^RA)^{-1}\cdot[e^R]=[A^{-1}]$
%    \item Compute $[\vek{w}]=[A^{-1}][\vek{b}]$ using secure multiplications and additions and open $\vek{w}$
%\end{enumerate}
%\vspace{-5pt}
%\caption{SIM}\label{prot:Inverse}
%\noindent\rule{\linewidth}{.8pt}\vspace{-\baselineskip}%
%\end{protocol}

\subsection{Communication Analysis}
We analyse the amount of information the parties need to send between each other depending on the size of the matrix in Protocol \ref{prot:regres}. We remark that the only it is only shares, multiplications, inversions and opening of shared values which require that parties need to send information. First party $i$ has to construct shares for $X_i$ and $\vek{z}_i$. This require him to send $(d^2+d)$ values to the $(n-1)$ other parties. Furthermore note that both multiplication and inversion also consists of openings. In worst case both a multiplication and an inversion costs 2 openings.\footnote{multiplication if both values needs to be converted to a degree-$t$ sharing and an inversion since it consists of a product where only the value $x$ we want to invert might need an conversion to degree-$t$ and then an opening of $xbe^r$}

%Hence, we count the number of these operations in the Gaussian approach and the inverse approach. There are many methods for computing matrix multiplication which asymptotically require less than $\mathcal{O}(d^3)$ but the constants hidden in the big-O notation might be huge. Thus, for this comparison we use the standard way to carry out matrix multiplication which requires exactly $d^3$ multiplications. Furthermore, SIM requires additional $d^2$ openings since we are opening the matrix $RA$. Afterwards we need to compute the product $[A^{-1}][\vek{b}]$ which requires $d^2$ multiplications and open $\vek{w}$ which gives $d$ openings. This gives a total of $d^3+d^2$ multiplications and $d^2+d$ additional openings.

For the Gaussian elimination part, we use $d-1$ inversions for computing $[\mathrm{inv}_k]$ and $\frac{d^2-d}{2}$ multiplications to compute $[\mathrm{frac}_{ik}]$. Then we use $\frac{d^3-d}{3}$ multiplications in the inner for-loop, and at last we do $d$ inversions, multiplications, and openings in the last for-loop. This gives a total of $\frac{2d^3+3d^2+d}{6}$ multiplications, $2d-1$ inversions, and additional $d$ openings. In conclusion, we get that the total amount of openings are at most $\frac{2d^3+3d^2+16d-6}{3}$.

%Since the information sent during the multiplication and inversion protocols are actually openings, we can collect these numbers to a total amount of openings. A multiplication consists in worst case of $2$ openings (if both shares are degree-$2t$ shares) and an inversion consists of a multiplication and an additional opening (the $xbe^r$ from Section \ref{sec:MPC_real}). Since $[be^r]$ is a degree-$t$ share we will in the worst case only have to convert $x$ before multiplying. Thus, in worst case an inversion costs $2$ openings.
\begin{table*}[t]
\resizebox{\linewidth}{!}{%
\begin{tabular}{|l|l|l|l|l|l|l|}
\hline
$\lambda$ &
  0.01 &
  0.1 &
  1 &
  10 &
  100 &
  1000 \\ \hline
\begin{tabular}[c]{@{}l@{}} GE \end{tabular} &
  0.17014 $\pm$ 0.02350 & 
  0.08442 $\pm$ 0.01782 & 
  0.03167 $\pm$ 0.00818 & 
  0.01737 $\pm$ 0.00384 & 
  0.01556 $\pm$ 0.00238 & 
  0.01614 $\pm$ 0.00224 \\ \hline
\begin{tabular}[c]{@{}l@{}} SGE ($n=3, t=1$)\end{tabular} &
  0.17014 $\pm$ 0.02350 & 
  0.08442 $\pm$ 0.01782 & 
  0.03167 $\pm$ 0.00818 & 
  0.01737 $\pm$ 0.00384 & 
  0.01556 $\pm$ 0.00238 & 
  0.01614 $\pm$ 0.00224 \\ \hline 
\begin{tabular}[c]{@{}l@{}} SGE ($n=5, t=2$)\end{tabular} &
  0.17014 $\pm$ 0.02727 &
  0.08442 $\pm$ 0.01782 &
  0.03167 $\pm$ 0.00818 &
  0.01737 $\pm$ 0.00384 &
  0.01556 $\pm$ 0.00238 &
  0.01614 $\pm$ 0.00224 \\ \hline
\begin{tabular}[c]{@{}l@{}} SGE ($n=7, t=3$)\end{tabular} &
  0.17014 $\pm$ 0.02350 &
  0.08442 $\pm$ 0.01782 &
  0.03167 $\pm$ 0.00817 &
  0.01737 $\pm$ 0.00384 &
  0.01556 $\pm$ 0.00238 &
  0.01615 $\pm$ 0.00224 \\ \hline
\begin{tabular}[c]{@{}l@{}} SGE ($n=9, t=4$)\end{tabular} &
  0.17016 $\pm$ 0.02351 &
  0.08442 $\pm$ 0.01782 & 
  0.03168 $\pm$ 0.00817 & 
  0.03370 $\pm$ 0.03224 &
  0.04106 $\pm$ 0.07312 &
  0.02682 $\pm$ 0.03297\\ \hline
\end{tabular}%
}
\caption{MSE results for different $\lambda$ with prior $\Sigma_{{w}} = I$ and $\vek{\mu}_{{w}} = \vek{0}$. We use security parameters $\sigma_{r}^2 = 10^8$, $\sigma_{\beta}^2 = 10^{8}$}
\label{tab:lambda_mse}
\end{table*}
% \usepackage{graphicx}
\begin{table*}[t]
\centering
\resizebox{\linewidth}{!}{%
\begin{tabular}{|l|l|l|l|l|l|l|l}
\cline{1-7}
$\lambda$ &
  0.01 &
  0.1 &
  1 &
  10 &
  100 &
  1000 &
   \\ \cline{1-7}
\begin{tabular}[c]{@{}l@{}}SGE ($\sigma_r^2=10^8, \sigma_\beta^2=10^{8}$)\end{tabular} &
  0.17014 $\pm$ 0.02727 &
  0.08442 $\pm$ 0.01782 &
  0.03167 $\pm$ 0.00818 &
  0.01737 $\pm$ 0.00384 &
  0.01556 $\pm$ 0.00238 &
  0.01614 $\pm$ 0.00224 &
   \\ \cline{1-7}
\begin{tabular}[c]{@{}l@{}}SGE ($\sigma_r^2=10^{10}, \sigma_\beta^2=10^{10}$)\end{tabular} &
  0.17014 $\pm$ 0.02350 &
  0.08442 $\pm$ 0.01781 &
  0.03167 $\pm$ 0.00818 & 
  0.01737 $\pm$ 0.00384 &
  0.01555 $\pm$ 0.00239 &
  0.01615 $\pm$ 0.00223 &
   \\ \cline{1-7}
\begin{tabular}[c]{@{}l@{}}SGE ($\sigma_r^2=10^{12}, \sigma_\beta^2=10^{12}$)\end{tabular} &
  0.17014 $\pm$ 0.02350 &
  0.08442 $\pm$ 0.01782 &
  0.03166 $\pm$ 0.00816 &
  0.01733 $\pm$ 0.00374 &
  0.01543 $\pm$ 0.00249 &
  0.01614 $\pm$ 0.00224 &
   \\ \cline{1-7}
\begin{tabular}[c]{@{}l@{}}SGE ($\sigma_r^2=10^{14}, \sigma_\beta^2=10^{14}$)\end{tabular} &
  0.17014 $\pm$ 0.02350 &
  0.08442 $\pm$ 0.01783 &
  0.03171 $\pm$ 0.00823 & 
  0.01751 $\pm$ 0.00390 & 
  0.39457 $\pm$ 0.86603 &
  0.45492 $\pm$ 1.18378 &
   \\ \cline{1-7}
\end{tabular}%
}
\caption{MSE results for different security parameters when $n=5$ and $t=2$. First row is the same row as $(n=5,t=2)$ in Table \ref{tab:lambda_mse}}
\label{tab:sigma_comp}
\end{table*}

% \usepackage{graphicx}
% \usepackage{graphicx}
\begin{table}[t]
\resizebox{\columnwidth}{!}{%
\begin{tabular}{|l|l|l|l|l|}
\hline
sample size ($m$): & 200 & 300 & 400 & 500 \\ \hline
\begin{tabular}[c]{@{}l@{}}$\sigma_\beta^2=10^{-6},\sigma_r^2=10^{-6}, \sigma_{e^r}^2=10^{-6}$, $\mu_{e^r}=1$\end{tabular} & 0.194 $\pm$ 0.132 & 0.040 $\pm$ 0.089 & 0.021 $\pm$ 0.034 & 0.013 $\pm$ 0.038  \\ \hline
\begin{tabular}[c]{@{}l@{}}$\sigma_\beta^2=100,\sigma_r^2=100, \sigma_{e^r}^2=16$, $\mu_{e^r}=10$
\end{tabular} & 0.233 $\pm$ 0.184 & 0.131 $\pm$ 0.147 & 0.014 $\pm$ 0.022  & 0.015 $\pm$ 0.031 \\ \hline
%\begin{tabular}[c]{@{}l@{}}$\sigma_\beta^2=10^6,\sigma_r^2=10^6, \sigma_{e^r}^2=100$, $\mu_{e^r}=20$ \end{tabular} & 0.458 $\pm$ 0.266 & 0.146 $\pm$ 0.192 & 0.065 $\pm$ 0.118 & 0.011 $\pm$ 0.030 \\ \hline
\begin{tabular}[c]{@{}l@{}}$\sigma_\beta^2=10^8,\sigma_r^2=10^8, \sigma_{e^r}^2=400$, $\mu_{e^r}=100$\end{tabular} & 0.147  $\pm$ 0.090 &  0.193  $\pm$ 0.167  &  0.103  $\pm$ 0.166  & 0.024  $\pm$ 0.018 \\ \hline
%\begin{tabular}[c]{@{}l@{}}$\sigma_\beta^2=10^{10},\sigma_r^2=10^{10}, \sigma_{e^r}^2=400$, $\mu_{e^r}=100$\end{tabular} &  0.359 $\pm$ 0.191 & 0.178 $\pm$ 0.208 & 0.103 $\pm$ 0.152 & 0.026 $\pm$ 0.050 \\ \hline
\begin{tabular}[c]{@{}l@{}}$\sigma_\beta^2=10^{12},\sigma_r^2=10^{12}, \sigma_{e^r}^2=400$, $\mu_{e^r}=100$\end{tabular} &  0.262   $\pm$ 0.230  & 0.246  $\pm$ 0.175 &  0.091 $\pm$ 0.153 &  0.026 $\pm$ 0.036 \\ \hline
\end{tabular}%
}
\caption{Mean plus/minus standard deviation of $p$-values for the $10$ conditional independence tests with $m$ samples.}
\label{tab:p_values}
\end{table}

\section{Experiments}

\subsection{Data set and expirimental setup}
In our experiments, we used Boston house price data set, a copy of UCI ML housing data set, that is contained in sklearn library \cite{Dua:2019,scikit-learn}, and which is suitable for linear regression. This data set contains 506 instances with 13 feature targeting the median value of the prices of the homes in \$1000’s. Before we applied our models to this data, we normalized it between $0$ and $1$.

\subsection{Testing accuracy}
In our experiments, we used the mean square error (MSE) to evaluate the accuracy of our secure regression method. To make our experiments trustworthy, we held 10 differently evaluations for each parameter combination. The data set is divided into train set (80\%) and test set (20\%). We shuffle the data set for each 10 evaluations. We divide the train set into $n$ subsets, one for each party, whenever we use secure computations for privacy preserving algorithms as mentioned in Section \ref{Sec_Notation}. Mean MSE results are reported across 10 different evaluations along with the corresponding standard deviations for the Gaussian elimination method with and without secure computation. 

%We used linear Support Vector Regression (SVR) \cite{chang2011libsvm}, which is also implemented using \cite{scikit-learn}, as a benchmark algorithm for comparison. The reason we used SVR is to have a robust linear regression method to compare with \cite{clarke2005analysis}. We tune the $C$ parameter of SVR from the set $\{0.01, 0.1, 1, 10, 100 \} $ using grid search.

In Table \ref{tab:lambda_mse}, we report the mean square error (MSE) for each algorithm using different $\lambda$ values from the set $\{0.01, 0.1, 1, 10, 100, 1000 \} $. We see that the secure version are very similar to the insecure one with respect to accuracy. However, we experience that increase in $t$ and $n$ may produce deviations in the MSE score for secure Gaussian elimination. For example, for SGE with $n = 9, t = 4$, we observe deviations for large $\lambda$'s. %Increase in $\lambda$ may also cause deviations in error metric.

In Table \ref{tab:sigma_comp}, we investigate the effect of the security parameters, $\sigma_r^2$ and $\sigma_\beta^2$, on the accuracy-privacy trade-off. Due to numerical problems, $e^r$ gave some $\mathrm{NaN}$ when $r$ was negative. Furthermore, very large $r$ was problematic here as well. Thus we have fixed $\mu_{e^r}=100$ and $\sigma_{e^r}^2=400$. We observe that as we increase $\sigma_r^2$ and $\sigma_\beta^2$, our secure computations become unstable and result in larger deviations from the right values. 

\subsection{Privacy analysis -- Boston data set}
We consider the privacy of our presented protocol in a $3$-party setting, where the computation is on the Boston data set. To analyze the privacy, we make use of the conditional independence test from \cite{GAN_independence} to argue that the conditional mutual information in \eqref{eq:Cond_MI} is small. We assume that the third party is corrupt and hence $V$ consists of; the randomness he chooses when secret sharing the entries in $X_3$ and $\vek{z}_3$, his shares of the preprocessed values for multiplications and divisions, the shares he receives for $X_i$ and $\vek{z}_i$, $i=1,2$, and the opened values during the protocol. We denote this $V$ by $V_3$ since it is the view of the third party and hence with this setup \eqref{eq:Cond_MI} becomes
\begin{align*}
    I(X_1,X_2,\vek{z}_1,\vek{z}_2;V_3|X_3,\vek{z}_3,\vek{w}).
\end{align*}
For simplicity and for illustration of how the security parameters impact the privacy we fixed $\alpha_i=i$ and $\mathcal{P}_s'=\{\alpha_1\}$, when generating the shares. We remark that intuitively it will be more secure to choose the elements in $\mathcal{P}_s'$ randomly but this seems to confuse the test to not reject the null hypothesis even with small security parameters when we do not have that many samples. So for illustration, we decided to fix $\mathcal{P}_s'$.

We want to test the null hypothesis that $X_1,X_2,\vek{z}_1,\vek{z}_2$ and $V_3$ are conditionally independent given $X_3,\vek{z}_3,\vek{w}$. Of course this is not the case since $V_3$ for instance includes shares for $X_1,X_2,\vek{z}_1,\vek{z}_2$, but with high security parameters, the dependency should be masked implying that it is close to be conditionally independent. Hence $I(X_1,X_2,\vek{z}_1,\vek{z}_2;V_3|X_3,\vek{z}_3,\vek{w})$ must be close to zero. To illustrate this, we apply the test from \cite{GAN_independence} and we need samples of $X_1,X_2,\vek{z}_1,\vek{z}_2,V_3,X_3,\vek{z}_3,\vek{w}$. We construct such samples in the following way. For a single sample, we choose randomly $20\%$ of the data instances from the Boston data set and split this subset in three, one for each party. From this subset we obtain a sample for $X_1,X_2,X_3,\vek{z}_1,\vek{z}_2,\vek{z}_3$. These are all the inputs to our secure protocol (Protocol \ref{prot:regres}), and running this protocol will give a sample for $V_3$ and $\vek{w}$. For some fixed security parameters ($\sigma_\beta^2,\sigma_r^2,\sigma_{e^r}^2$) this process is repeated $2000$ times to obtain $2000$ samples of the random vectors. We feed the conditional independence test $10$ times with $m$ randomly chosen of these samples. We report the mean and standard deviation of the $p$-value in Table \ref{tab:p_values}. As we observe, the conditional independence test needs more samples before it can determine any conditional dependencies, and reject the null hypothesis, when we are increasing the security parameters. 

\section{Conclusion and future work}
We have described a secure protocol for obtaining a linear model based on a prior and on a number of data sets distributed between several parties. We have illustrated that carrying out the computations using MPC techniques based on real number secret sharing we can obtain the model and argued that this technique is secure even when multiple secure operations are carried out sequentially. To argue security we showed that by increasing the security parameters it is hard to determine any dependency between the information we want to hide and the messages received during the protocol. However, if the security parameters become too high, we observed some inaccuracy in the secure Gaussian elimination protocol due to numerical problems. In this paper, we have focused on how real number MPC can be used to obtain a linear model on a split data set but it could be interested to consider  other models as well. However, if the model can be obtained from linear combinations, products and inversions we have in this paper the tools to obtain the model even though further analysis of the privacy is of course needed. Furthermore, it could be interesting to consider other secure operations to expand the toolbox for secure computations. 

\section*{References}
\printbibliography[heading=none]

\end{document}